\documentclass[10pt,twocolumn,letterpaper]{article}

\usepackage{wacv}              

\usepackage{xcolor}
\usepackage{cite}
\usepackage{amsmath,amssymb,amsfonts}
\usepackage{algorithmic}
\usepackage{graphicx}
\usepackage{textcomp}
\usepackage{adjustbox}
\usepackage{booktabs}
\usepackage{hyperref}
\usepackage{caption}
\usepackage{soul}
\usepackage{placeins}
\usepackage{float}
\usepackage{siunitx}

\usepackage[capitalize]{cleveref}
\crefname{section}{Sec.}{Secs.}
\Crefname{section}{Section}{Sections}
\Crefname{table}{Table}{Tables}
\crefname{table}{Tab.}{Tabs.}

\def\wacvPaperID{1712} 
\def\confName{WACV}
\def\confYear{2025}

\begin{document}

\title{BASED: Bundle-Adjusting Surgical Endoscopic Dynamic Video Reconstruction using Neural Radiance Fields}


\author{
Shreya Saha$^1$ \hspace{0.7cm}
Zekai Liang$^1$ \hspace{0.7cm}
Shan Lin$^1$ \hspace{0.7cm}
Jingpei Lu$^1$ \hspace{0.7cm}
Michael Yip$^1$ \hspace{0.7cm}
Sainan Liu$^2$ \\[6pt]
$^1$University of California, San Diego, USA \hspace{1.5cm}  $^2$Intel Labs, USA\\
{\tt\small \{ssaha, z9liang, shl102, jil360, yip\}@ucsd.edu}\hspace{0.6cm} {\tt\small sainan.liu@intel.com}
}

\maketitle
\begin{abstract}
   Reconstruction of deformable scenes from endoscopic videos is important for many applications such as intraoperative navigation, surgical visual perception, and robotic surgery. It is a foundational requirement for realizing autonomous robotic interventions for minimally invasive surgery. However, previous approaches in this domain have been limited by their modular nature and are confined to specific camera and scene settings. Our work adopts the Neural Radiance Fields (NeRF) approach to learning 3D implicit representations of scenes that are \textit{both dynamic and deformable} over time, and furthermore with unknown camera poses. This work removes the constraints of known camera poses and overcomes the drawbacks of the state-of-the-art unstructured dynamic scene reconstruction technique, which relies on the static part of the scene for accurate reconstruction. Through several experimental datasets, we demonstrate the versatility of our proposed model to adapt to diverse camera and scene settings, and show its promise for both current and future robotic surgical systems. 
\end{abstract}

\section{Introduction}
\label{sec:intro}
Surgical scene reconstruction from endoscopic videos is an important and challenging task in robotic-assisted minimally invasive surgery~\cite{yip2023artificial}. It also has widespread use in many downstream clinical applications such as surgical workflow analysis, image-guided robotic surgery, automation, surgical environment simulation, and augmented reality to train surgeons to be familiar with preoperative assessment and planning. Many learning and automation algorithms for robotic surgery are reliant on accurate 3D reconstruction of the surgical scenes~\cite{liu2021real, liang2023real, shinde2024jiggle}.

The majority of current approaches for reconstructing surgical scenes from endoscopic videos adopt a modular sequence of tasks, each tackled by independent models~\cite{long2021dssr,recasens2021endo,Li_2021_ICCV,lurie2017comprehensive,soper2012bladder,hu2012robust,widya2019sfm,Phan2020,widya20193d_jtehm}. This fragmented methodology is susceptible to accumulating errors. For instance, a significant number of existing methods either estimate or rely on ground truth camera poses before proceeding with scene reconstruction. Errors in pose estimation often lead to faulty scene reconstructions. Camera poses obtained through structure from motion (SfM) or simultaneous localization and mapping (SLAM)~\cite{lamarca2020defslam, rodriguez2021sd,song2018misslam,garciagrasa2013visualslam,liu2022sage} often exhibit noise. When the camera is fixed to the end effector of a robotic arm, camera pose estimation can be accomplished through forward kinematics. But it's prone to errors stemming from in joint angle measurements inaccuracies attributed to instrument sag, cable stretch, positioning biases, and reading drift~\cite{richter2021robotic}. Furthermore, in addition to being modular, these methods will not generalize from static to deformable scene settings.

\begin{figure}[t]
\centering
\small
Ground Truth Color (left) and Depth (right)\\
\vspace{1mm}
\includegraphics[width=\linewidth,clip=true,trim={2mm 2mm 2mm 2mm}]{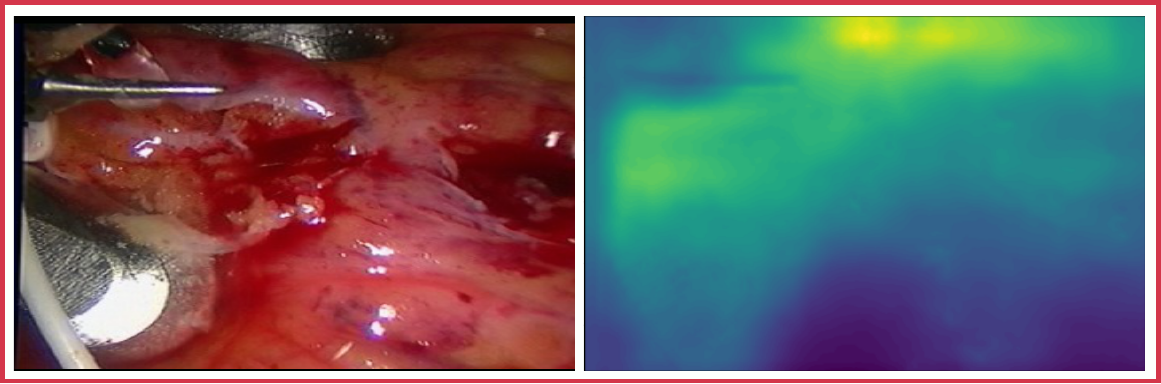}\\
RoDynRF (SOTA) color and depth\\
\vspace{1mm}
\includegraphics[width=\linewidth,clip=true,trim={2mm 2mm 2mm 2mm}]{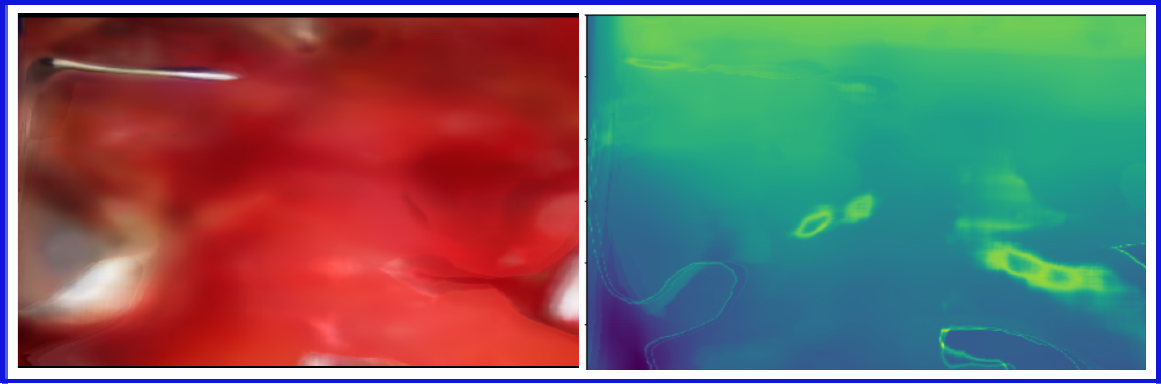}
\textbf{BASED (Ours) color and depth}\\
\vspace{1mm}
\includegraphics[width=\linewidth]{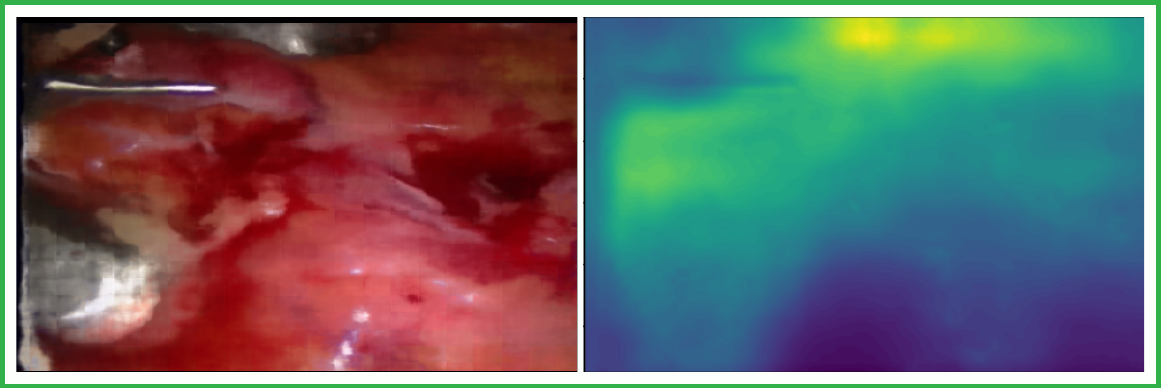}
\vspace{-4mm}
\captionsetup{font=footnotesize}
\caption{BASED is a novel NeRF-based method that can be used in dynamic and deformable scenes with unknown camera poses. It can produce novel viewpoint renderings  with robust color (left) and depth (right) reconstructions, even from monocular untracked camera images. Comparisons with state-of-the-art methods show that it has leading performance in scene reconstruction. Depth in the ground truth row is the reference depth, estimated from \cite{recasens2021endo}.}
\label{fig:teaser}
\vspace{-5mm}
\end{figure}


Neural Radiance Fields (NeRF)~\cite{mildenhall2020nerf} has shown tremendous performance in the domain of 3D reconstruction. They have been applied recently to stereo-endoscopic videos for deformable surgery scene reconstruction~\cite{wang2022endonerf,zha2023endosurf}. However, some constraints in these methods are that the camera should be static with known stereo camera information. 
Alternatively, some works have extended NeRF with camera pose optimization (e.g., \cite{chen23dbarf,liu2023baangp}). 
However, these methods generally assume the scene to be static. 


More recently, 3D Gaussian Splatting (3DGS)~\cite{kerbl20233d} represents the scene with a cloud of 3D Gaussians and estimate the Gaussian parameters based on a efficient rendering algorithm. Similar to NeRF, this technique has been applied to deformable surgical scenes with fixed or known camera poses~\cite{zhuendogs,liu2024endogaussian}, as well as simultaneous estimation of static scene representations and unknown camera poses \cite{fu2023colmap}. 
Although 3DGS has achieved state-of-the-art visual rendering quality, it does not capture the 3D geometric information of the scene accurately enough since surgical videos are typically highly under-constrained where the camera movement is relatively small. Moreover, since training of 3DGS primarily focuses on minimizing the difference between rendered and actual images rather than learning accurate 3D geometry of the scene, the geometric accuracy of the Gaussians highly depends on the quality of the point cloud and camera poses initialized using off-the-shelf SfM methods like COLMAP~\mbox{\cite{schoenberger2016sfm, schoenberger2016mvs}}, which still struggle to handle certain surgical videos. Therefore, we develop our method based on NeRF and demonstrate its improved ability to capture scene geometry through several experiments. 

In summary, we present the first NeRF-based method that can reconstruct deformable surgical scenes captured from untracked monocular endoscopic videos. The overall method combines a learnable pose layer that parameterizes camera poses, a deformation module that learns the deformation of a 3D point at a given time step with respect to a canonical position and a rendering module that takes the canonical coordinates of a 3D point at every time step along with the 2D camera directions and outputs the volumetric density and color information. 
We can also capture more accurate geometric information without compromising on rendering capability for deforming scenes, even in instances where the camera motion is unknown. 
This makes the method applicable to a wide range of robotic systems and semi-autonomous applications.
The major contributions of this paper are thus:
\begin{itemize}
    \item A NeRF approach for \textit{dynamic and deformable} scenes from data captured at \textit{unknown camera poses},
    \item A method for describing multi-view correspondence loss, specifically catered to a dynamic and deformable setting, that enables establishing correspondence across spatial temporal drift in tissues, and 
    \item A framework that offers a generic, unrestricted, and versatile solution to robotic surgical perception that is validated on several surgical scene reconstruction datasets. 
\end{itemize}

\section{Related Work}

\subsection{Traditional Scene Reconstruction Techniques} 

Structure from Motion (SfM) utilizes a series of two-dimensional images to construct a three-dimensional model of a scene or an object, and it is widely used in the domain of scene/organ/tissue reconstruction with endoscopic images~\cite{lurie2017comprehensive, soper2012bladder, hu2012robust, widya2019sfm, Phan2020, widya20193d_jtehm}. However, SfM techniques like COLMAP~\cite{schoenberger2016sfm, schoenberger2016mvs} often fail to estimate camera poses for the entire length of sequences. SLAM-based approaches~\cite{garciagrasa2013visualslam, song2018misslam,lamarca2020defslam, rodriguez2021sd, liu2022sage} have been employed in the field of endoscopic scene reconstructions to model sparse and dense tissue surfaces. The scene is usually assumed to be static, and the estimated deformations do not always look natural.

\subsection{Static and Dynamic NeRF} 
NeRFs \cite{mildenhall2020nerf} are implicit volumetric representations that encode the appearance and geometry of a 3D scene. It learns a continuous representation of the scene, enabling novel view renderings, unlike SLAM/SfM techniques that usually learn a discrete representation. It takes a 3D point coordinate and a 2D camera viewing direction and outputs RGB color information and the density of the point. The original NeRF model strictly catered to static scenes. D-NeRF~\cite{pumarola21dnerf} is the first work that expanded the NeRF model to learn deformable scenes. It consists of two separate models that progressively learn the deformation and the volumetric scene reconstruction simultaneously. Robust Dynamic NeRF, or RoDynRF~\cite{liu2023robust}, tries to reconstruct dynamic scenes by estimating the camera poses using only the static part of the scene, which is not suitable for surgical scene reconstruction given that the scenes are highly deformable. 
EndoNeRF~\cite{wang2022endonerf} builds on D-NeRF for endoscopic scenes by adding a tool-guided ray casting 
and by sampling points along a ray closer to the tissue surfaces using a Gaussian transfer function but relying on stereo depth. EndoSurf~\cite{zha2023endosurf} further improved upon EndoNeRF~\cite{wang2022endonerf} by employing three networks to learn the deformation, signed distance function (SDF), and color of a viewpoint. However, as seen in Figure~\ref{fig:teaser}, these methods strictly cater to cases where the camera is static. If extended to moving camera sequences with camera poses estimated by COLMAP, the rendering quality degrades significantly. 

\subsection{Static and Dynamic Gaussian Splatting}
3DGS is a fast rendering technique for novel view synthesis. It uses 3D Gaussians for scene representation and a novel tile-based rasterizer for fast 3D Gaussian rendering. Given a set of camera poses and their corresponding 2D views, a cloud of 3D Gaussian representation can be learned end-to-end. While the original work focused on static scenes, dynamic and deformable scenes have also been explored in recent months.  

Dynamic 3D Gaussians \mbox{\cite{luiten2023dynamic}} builds a static scene using 3DGS for the first frame, then incrementally learns the offset for each 3D Gaussian. More recently, 3DGStream \mbox{\cite{sun20243dgstream}} has improved the efficiency of offset cache storage and enabled new Gaussian spawning. However, these models can only work with multiview images and rely heavily on the information from the first frame. 

Yang \textit{et al.} \mbox{\cite{yang2023deformable3dgs}} and Wu \textit{et al.} \mbox{\cite{wu20234dgaussians}} both use a deformation field to model offsets to a canonical space of explicit 3D Gaussians, whereas \mbox{\cite{yang2023gs4d}} has increased GS dimension into 4D. However, these methods assume ground truth camera poses are known. Our model does not have this assumption.

\subsection{Joint Pose Estimation and Scene Reconstruction}
Bundle adjusting neural radiance fields (BARF)~\cite{lin2021barf} first proposed an end-to-end NeRF-based framework that can jointly estimate camera extrinsics and reconstruct the 3D scene. It implements a coarse-to-fine technique to gradually activate the higher frequency components of the positional encoding scheme that aids in pose refinement,  and only uses the photometric loss for backpropagation through the network. There have been numerous follow-up works (~\cite{liu2023baangp, jeong21scnerf, truong23sparf, chng2022gaussian, cheng2023lunerf, chen23dbarf, xia22sinerf, bian2022nopenerf, meuleman2023localrf}) that improved BARF further. However, all these methods specifically deal with static scenes. RoDynRF has firstly implemented estimating camera poses while simultaneously reconstructing dynamic scenes. However, RoDynRF requires the dynamic pixels to be masked out prior to training. This is hard to acquire as the entire surgical scene may be deformable and textureless. Hence, as seen in Figure \ref{fig:teaser}, it cannot be generalized for endoscopic videos.
Our proposed model overcomes the limitations of EndoNeRF and EndoSurf by expanding to moving cameras with unknown poses. It does not constrain part of the scene to be static for it to be able to learn deformations, thus also overcoming the limitations of RoDynRF.

\begin{figure*}[htb!]
\centering
\includegraphics[width=\linewidth]{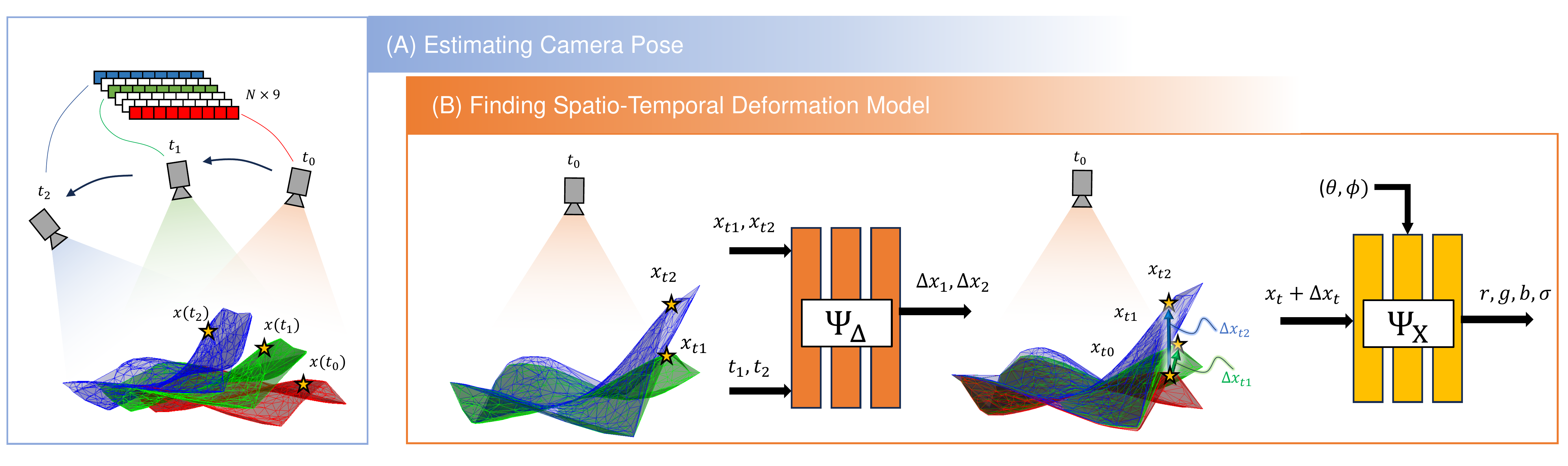}
\vspace{-8mm}%
\captionsetup{font=footnotesize}
\caption{Method overview: (A) The first part of the diagram shows a tissue that appears to be dynamic and deformable at three timestamps $t_0$, $t_1$, and $t_2$. Camera poses are estimated by a $N\times9$ learnable matrix through the camera pose estimation layer. $x(t_0), x(t_1)$ and $x(t_2)$ shows the trajectory taken by the same point $x$ on the tissue at these three different timesteps. (B) $x_{t_1}$ and $x_{t_2}$ represents non-rigid deformation trajectory of $x$ at time $t_1$ and $t_2$. The deformation model denoted by$\psi_{\Delta}$ takes in $x_{t_1}$ and $x_{t_2}$ and predicts their displacement from the canonical configuration of the scene. The canonical model $\psi_x$ then takes in the mapped canonical position of the point $x_{t_0} = x_{t_1} + \Delta x_{t_1} = x_{t_2} + \Delta x_{t_2}$ along with 2D camera directions, and predicts the color and density information.}
\label{fig:overview}
\vspace{-4mm}
\end{figure*}

\section{Methodology}

Figure \ref{fig:overview} gives a high-level overview of our method. The overall model consists of three parts: (1) the pose estimation module~(see \ref{sec:camera_module}), (2) the deformable module~(see \ref{sec:deformation_module}), and (3) a canonical NeRF module~(see \ref{sec:canonical_module}). 


\subsection{Camera Pose Module}
\label{sec:camera_module}
We use a camera pose module to model the appearance of the rigid-body motion of the scene as is shown in Figure~\ref{fig:overview} (A). We follow the convention of previous work~\cite{lin2021barf} and model the camera pose with $P_i = [R_i| t_i] \in SE(3)$ representing camera-to-world transform of camera $i$, where $R_i \in SO(3)$ denotes the rotation and $t_i \in\mathbb{R}^3$ denotes the translation. We use $K\in\mathbb{R}^{3\times3}$ to represent the intrinsic matrix.

In terms of implementation details, the poses are parameterized as a learnable parameter $P_i \in \mathbb{R}^9$ for $i \in [1, N]$ for $N$ frames. The first six columns of the matrix refer to the first two rows of the camera rotation matrix, and the last three values refer to the camera translation vector. The third row of the rotation matrix can be estimated by taking a cross-product of the other two rows. The pose layer is initialized to identity and updated for the first 200 iterations, after which the layer is frozen.

\subsection{Deformation Module}
\label{sec:deformation_module}
After the camera pose module explains away the rigid-body motion, the deformation module will model the non-rigid-body deformation of the scene from a canonical state at $t_0$ as is shown in Figure~\ref{fig:overview} (B).
The deformation module $\psi_\Delta(x,t)$ takes as input 3D spatial coordinates $x\in\mathbb{R}^3$ and a time step $t$ and learns the deformation $\Delta x$ of the coordinate $x$ from time step zero: 
\begin{equation}
\psi_{\Delta}(x,t) = 
\begin{cases}
        \Delta x  & \text{if } t \neq 0\\
        0 & \text{if } t = 0
    \end{cases}
\end{equation}
If $x_{0}$ is the coordinate of a point at time $t=0$ and it deforms to $x_t$ at time t, then we can say that
$    \psi_{\Delta}(x_t, t) + x_t = x_0$.

\subsection{Canonical NeRF Module}
\label{sec:canonical_module}
This module will capture the implicit 3D representation of the canonical state of the scene.
The canonical module $\psi_x$ can be assumed to be a vanilla NeRF model, which takes as input a spatial 3D location and 2D camera orientations, and outputs color (RGB values) and the volumetric density of the location. $x_0$ calculated above is the input to the canonical model along with camera viewing directions $d$. The rendered output of the canonical model (color $c$, and density $\sigma$) is corresponding to the deformed point $x_t$ at time $t$.
\begin{equation}
\psi_x(x_t + \psi_{\Delta}(x_t,t), d) = c ,\sigma    
\end{equation}

We apply conventional positional encoding~\cite{pumarola21dnerf} to all the inputs, $<(sin(2^l \pi p), cos(2^l \pi p))>_0^L$, with $L=10$ for 3D point coordinates $x$ and time frame $t$, and $L=4$ for camera directions $d$.


\subsection{Tool mask-guided ray casting and Stereo Depth Cue Ray Marching}

Surgical tools occlude part of the scene in most of our sequences. Instead of shooting rays randomly all through the image, we adopt a tool mask-guided ray casting strategy~\cite{wang2022endonerf}. Using binary tool masks, an importance map is created which highlights the regions containing tool pixels. During training, instead of random ray sampling throughout the image, we only sample pixels from this distribution using inverse transform sampling. Secondly, we sample pixels close to the tissue surface using a Gaussian transfer function and use a depth refinement strategy to get rid of corrupt artifacts in the estimated depths.

\subsection{Losses}
\begin{figure}
\centering
\includegraphics[width=0.9\linewidth,clip=true,trim={0 0 0 12mm}]{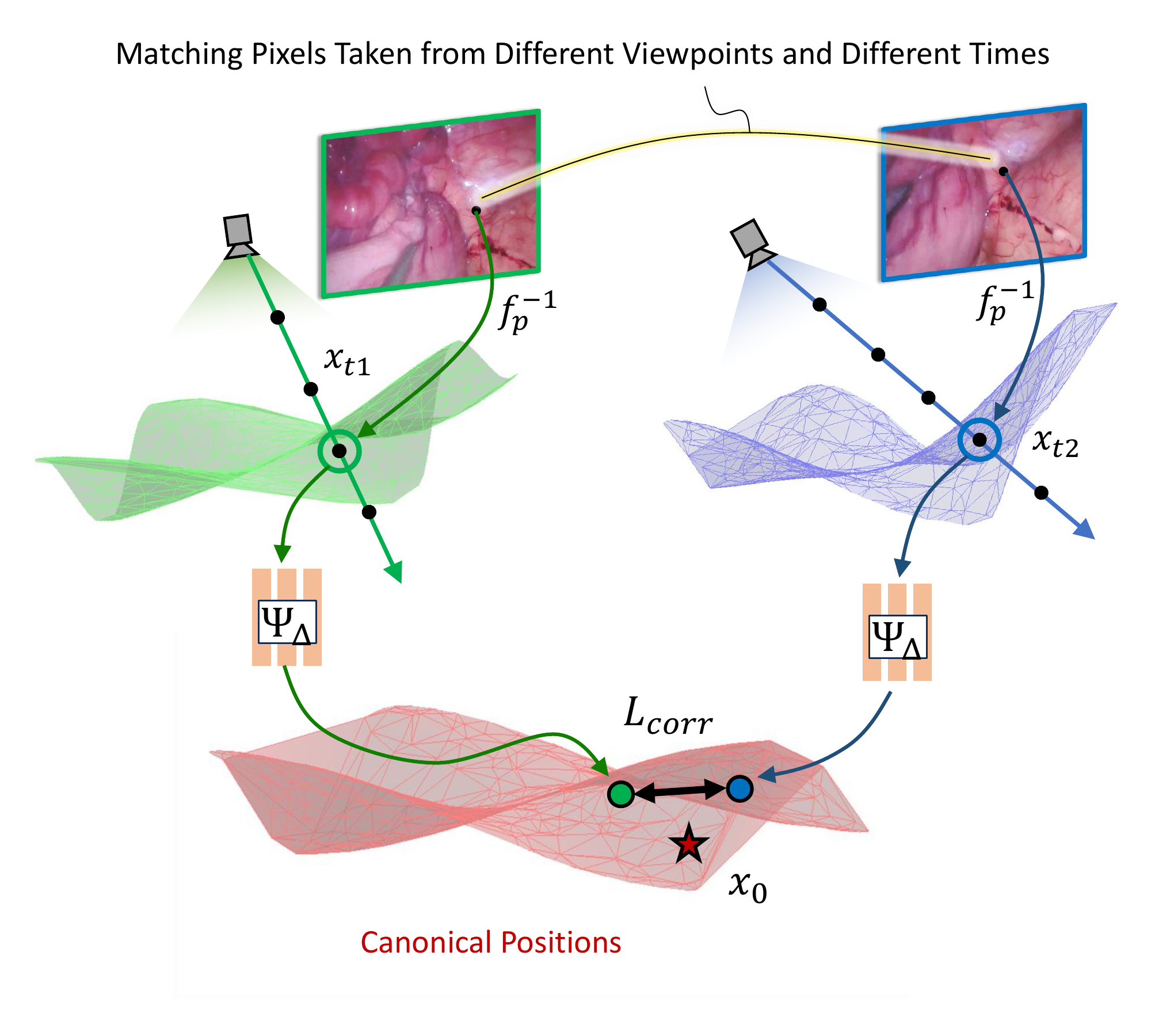}
\captionsetup{font=footnotesize}
\caption{Overview of the correspondence loss: Matching pixels are chosen from a random pair of images (having correspondence above a certain threshold) using PDC-Net \cite{truong2021learning}. The matching pixels are projected from 2D pixel space to their 3D positions in the world frame, and then further passed through the deformation model to get their canonical 3D positions. The difference between the mapped 3D points is used to calculate the flow correspondence loss. The loss is designed to pull the outputs of correspondence points returned by the deformation model closer together towards the same canonical point denoted as $x_0$.}
\label{fig:flowloss}
\end{figure}
Our model uses photometric loss, correspondence loss, and estimated depth Loss for monocular dynamic camera deformable scene reconstruction.


\begin{table*}[htbp]
\caption{Ablation studies on Hamlyn Rectified18-1 dataset with dynamic motions.  $\checkmark$ indicates loss usage. BA stands for bundle-adjusting. EndoNeRF experiment is using COLMAP for camera pose estimation.}
    \centering
    \begin{adjustbox}{width=\linewidth,center}
    \small
    \begin{tabular}{@{\extracolsep{35pt}}c||ccc|ccc}
    \hline
         &BA& $L_d$ &$L_{corr}$ &  PSNR$\uparrow$&SSIM$\uparrow$&LPIPS$\downarrow$\\
         \hline
         EndoNeRF~\cite{wang2022endonerf}&COLMAP&\checkmark&-&26.97&0.895&0.164\\
         \hline
         BASED w/o $L_dL_{corr}$&\checkmark&-&-&30.236&0.925&0.124\\
         BASED w/o $L_d$&\checkmark&-&\checkmark&31.415&0.935&0.108\\
         BASED w/o $L_{corr}$ &\checkmark&\checkmark&-&31.403&0.937&\textbf{0.102}\\
         BASED (final) &\checkmark&\checkmark&\checkmark&\textbf{32.227}&\textbf{0.944}&0.106\\
         \hline
    \end{tabular}
    \label{tab:rectified18_1_ablation}
    \end{adjustbox}
     \begin{flushleft}
  
    \end{flushleft}
\end{table*}

    \emph{\textbf{RGB Photometric Loss}}:
    We use photometric loss based on conventional setup in NeRF models~\cite{mildenhall2020nerf}. At every training iteration, we compare the rendered color of the sampled pixels with the ground truth colors using the $L_2$ loss, and this loss is backpropagated through the entire network with all three modules.
    \begin{equation}
        L_{pho} = \frac{1}{N} \Sigma_{i=1}^N || \hat{C_{i}}(p,t) - C_{i}(p,t) ||_2^2
    \end{equation}
    where $\hat{C_{i}}$ is the rendered color and $C_{i}$ is the ground truth color of the pixel $p$ at time $t$ with respect to frame $i, ~i=0...N$.

    \emph{\textbf{Dynamic Multi-View Correspondence Loss}}:
    We introduce a novel 3D multi-view correspondence loss for dynamic scenes to offer extra constraints on pose estimation. Figure \ref{fig:flowloss} gives a high-level idea of this loss. For a given pair of images, $I_i^a$ (viewpoint $P_i$ at time $t_a$) and $I_j^b$ (viewpoint $P_j$ at time $t_b$) with matching pixels $p^a \in I_{i}^a$ and $q^b \in I_j^b$, both pixels $p^a$ and $q^b$ should project to the same point in 3D space in the canonical space. $z_p^a$ and $z_q^b$ are the estimated depths of the pixels $p^a$ and $q^b$ respectively. Let $f_P^{-1}$ be the function that maps the 3D coordinate of a pixel given its estimated depth and its respective camera viewpoint. Let $x_p^a \in\mathbb{R}^3$ be the 3D coordinate of pixel $p^a$ at time $t_a$ and $x_q^b \in\mathbb{R}^3$ be the 3D coordinate of pixel $q^b$ at time $t_b$. Then, the estimated coordinates are
    \begin{equation}
        x_{p}^a = f_{P_i}^{-1} (p^a, z_p^a)
    \end{equation}
    \begin{equation}
        x_{q}^b = f_{P_j}^{-1} (q^b, z_q^b)
    \end{equation}
    If there was no deformation, $x_{p}^a$ and $x_{q}^b$ should be the same point in 3D space. However, since this is a deformable scene, we need to map both the points back to their canonical state in order to calculate the loss:
    \begin{equation}
        \hat{x}_{p}^a = \psi_\Delta(x_p^a,t_a) + x_p^a
    \end{equation}
    \begin{equation}
        \hat{x}_{q}^b = \psi_\Delta(x_{q}^b,t_a) + x_q^b
    \end{equation}

Then, the loss becomes:
     \begin{equation}
        L_{corr} = w (L_\delta(\hat{x}_{p}^a , \hat{x}_{q}^b))
    \end{equation}
where $w$ is loss weight term and $L_\delta$ is the Huber loss function. 

    \emph{ \textbf{Estimated Depth Loss}}:
In spite of only having 2D images, we can use an \textit{estimated} depth loss to further assist better reconstruction when the camera pose estimation is optimized. The depth estimates can be found via an off-the-shelf image-to-depth estimator.
    This loss is only used to optimize the deformable and canonical modules after the pose layer is frozen. Applying the estimated depth loss before the camera pose matrix is optimized leads to abrupt artifacts. We collect depths estimated from RGB images via existing pre-trained methods~\cite{sttrlight},~\cite{recasens2021endo} during training. We compare the rendered depth values of the sampled pixels with the corresponding estimated depth values using a $L_2$ loss. The gradient from this loss is only backpropagated through the deformable NerRF modules, after the pose layer is frozen.
    \begin{equation}
        L_{d} = \frac{1}{N} \Sigma_{i=1}^N || \hat{D_{i}}(p,t) - D_{i}(p,t) ||_2^2
    \end{equation}
    where $\hat{D_{i}} (p,t) = \frac{1}{\sigma_i(p, t)}$ is the rendered depth, $\sigma_i$ is the rendered density and $D_{i}(p, t)$ is the reference depth of the pixel $p$ at time $t$ with respect to frame $i \in [1, N]$ for $N$ frames.
Note that if a hardware solution to depth is available (e.g., stereo cameras or RGB-D sensor, the ground truth depth can be used as well to calculate an estimated depth loss.

\begin{figure}[h]
\centering
\footnotesize
\begin{tabular}{cc}
EndoNeRF ~~~~~~~~~~~~~&~~~~~~~~~~~~~ BASED \\
\end{tabular}
\includegraphics[scale=0.45]{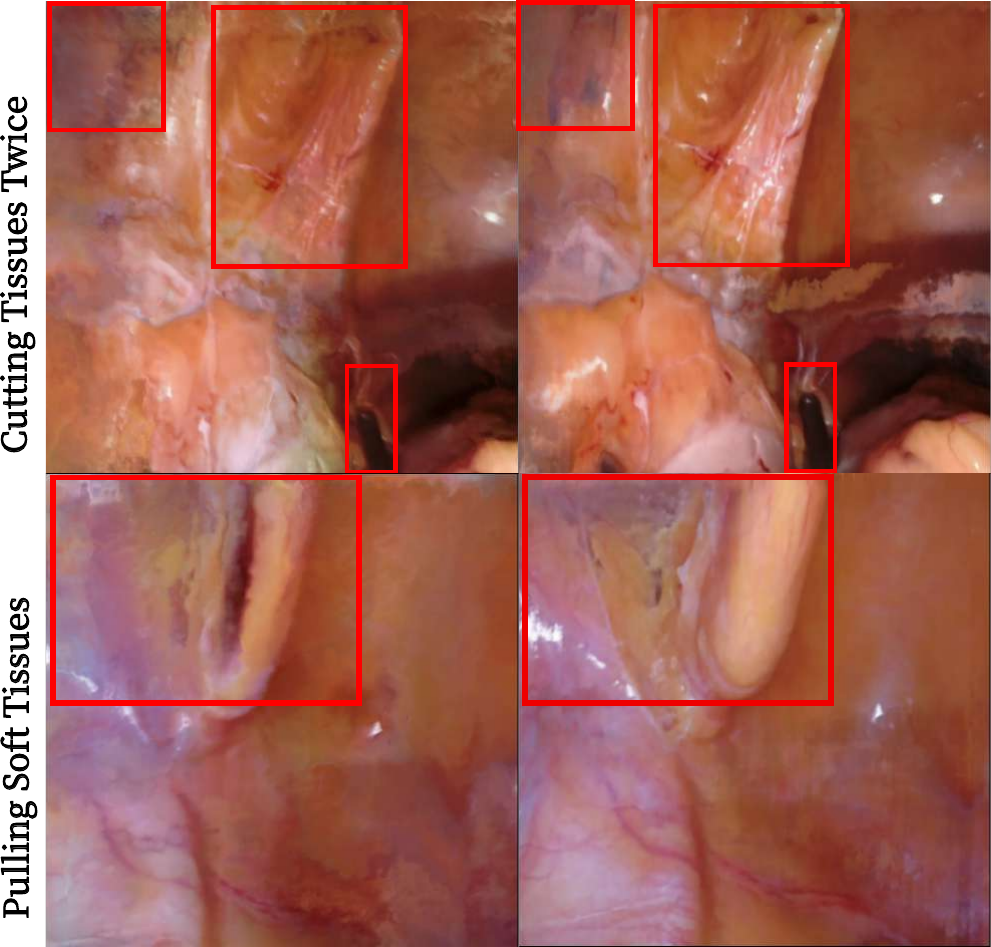}
\captionsetup{font=footnotesize}
\caption{Qualitative analysis of our proposed model BASED 
and EndoNeRF on "Cutting Tissues Twice" dataset for novel view renderings. Highlighted sections show how BASED is able to avoid different artifacts in the images, resulting in a sharper image.}
\label{fig:novel_view}
\vspace{-5mm}
\end{figure}

\section{Experiments}

\subsection{Implementation Details}
All of the experiments have been performed on a single Nvidia RTX-A6000 GPU. The weights of the NeRF network and the pose estimation network are simultaneously updated with a learning rate of 5e-4. After 200 iterations, the pose layer is fixed, and only the NeRF model is updated for another 100K iterations. In the first stage, one batch consists of all of the training images, because it requires more views of the scene for better pose estimation. In the second stage, one batch consists of a single training image for faster training. Tool masks are manually created for the Hamlyn dataset sequences, partly using Segment Anything Model~\cite{kirillov2023segany}. We use an off-the-shelf dense pixel correspondence network, PDCNet~\cite{truong21dense}, to obtain corresponding matching pixels between a pair of images for calculating the multi-view correspondence loss $L_{corr}$.

\begin{figure*}
\scriptsize
\begin{tabular}{p{0.25in}p{1.0in}p{1.2in}p{1.2in}p{1.2in}p{1.2in}}
& Ground Truth & BASED w/o $L_dL_{corr}$ & BASED w/o $L_d$ & BASED w/o $L_{corr}$ & BASED (final)
\end{tabular}
\centering
\includegraphics[width=\linewidth]{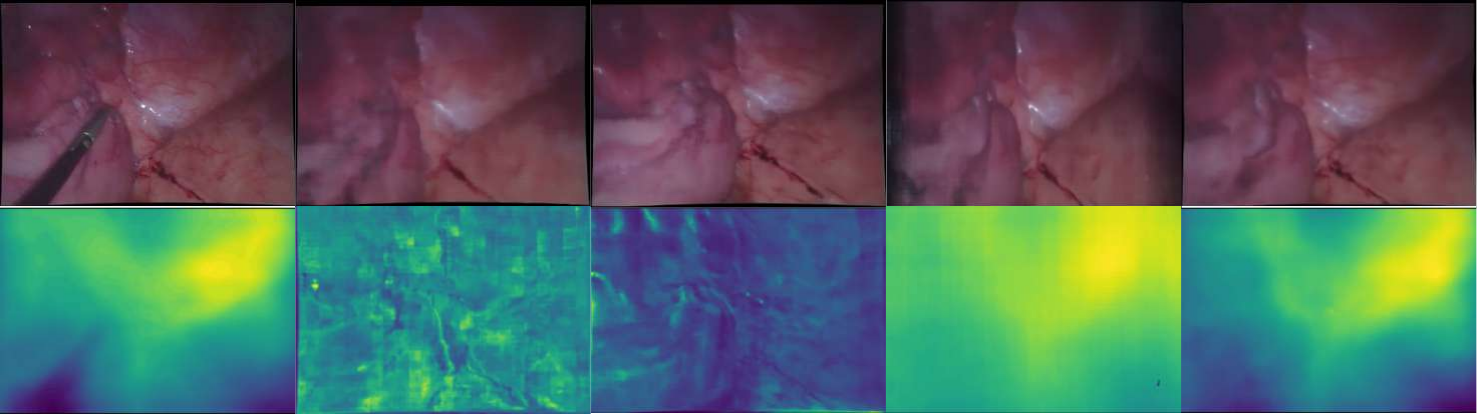}
\vspace{-3mm}
\captionsetup{font=footnotesize}
\caption{Ablation analysis on Hamlyn Rectified 18-1 dataset shows how different losses bring about an improvement in the rendered results from the final model. Please note that the depth in the Ground Truth column is the reference depth.}
\label{fig:depth_ablation}
\end{figure*}

\begin{figure*}[ht]
\centering
\footnotesize
\includegraphics[width=0.97\linewidth, clip=true, trim={0 0 0 0}]{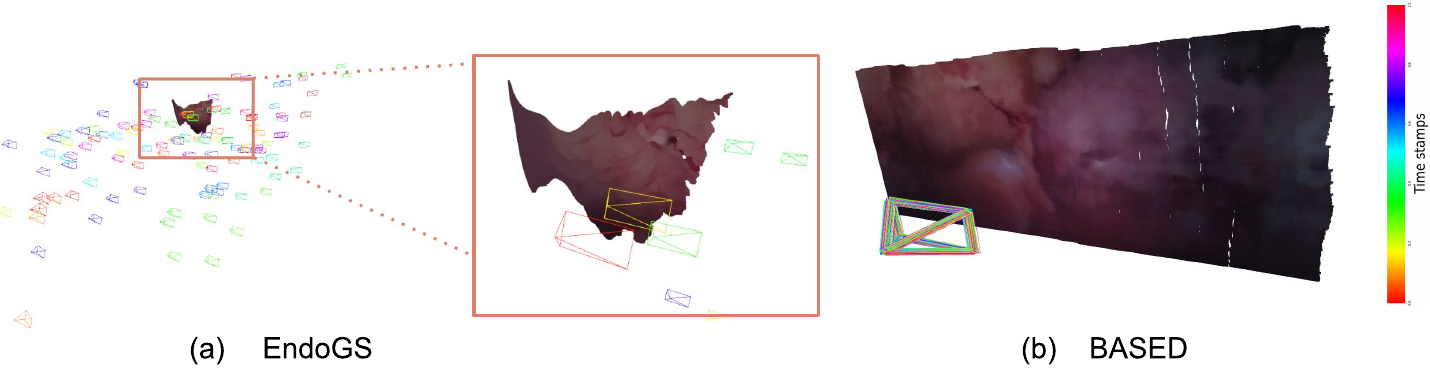}
\vspace{2mm}
\captionsetup{font=footnotesize}
\vspace{-3mm}
\caption{Camera poses and point cloud visualization of (a) EndoGS and (b) BASED on hamlyn dataset Rectified18-2. The colors assigned to the cameras correspond to their timestamps in the video.}
\label{fig:based model}
\end{figure*}

\begin{table*}[h]
 \centering
    \caption{Quantitative analysis of BASED and State-of-the-Art on the Hamlyn Dataset.} 
    \label{tab:quantitative_analysis}
    \begin{adjustbox}{width=\linewidth,center}
    \small
    \begin{tabular}{l||cccc|cccc|cccc}
        \toprule
        & \multicolumn{4}{c}{PSNR $\uparrow$} & \multicolumn{4}{c}{SSIM $\uparrow$} & \multicolumn{4}{c}{LPIPS $\downarrow$}\\
        \cmidrule(r){2-13}
        &EndoNeRF&RoDynRF &EndoGS&Ours&EndoNeRF&RoDynRF &EndoGS&Ours&EndoNeRF&RoDynRF&EndoGS&Ours\\
        \midrule
         Rectified04-1 (scds)&-&\underline{14.902}&-&\textbf{20.947}&-&\underline{0.468}&-&\textbf{0.621}&-&\underline{0.577}&-&\textbf{0.269}\\
         Rectified18-1 (mcds)&22.838&25.700&\underline{30.821}&\textbf{32.277}&0.817&0.874&\underline{0.928}&\textbf{0.944}&0.273&0.192&\textbf{0.100}&\underline{0.106}\\
         Rectified18-2 (mcds)&29.572&24.690&\textbf{35.969}&\underline{33.017}&0.869&0.875&\textbf{0.956}&\underline{0.936}&0.235&0.201&\textbf{0.079}&\underline{0.103}\\
         \bottomrule
    \end{tabular}
  \end{adjustbox}
 \begin{flushleft}
`-' denotes that the model fails to complete the task.\\
Best results are \textbf{bolded}; second best are \underline{underlined}.
\end{flushleft}
\end{table*}

\begin{table*}[h]
    \centering
    \caption{Quantitative analysis of BASED and state-of-the-art on the EndoNeRF da Vinci dataset.}
    \label{tab:endonerf_reuslts}
    \begin{adjustbox}{width=\linewidth,center}
    \small
    \begin{tabular}{l||cccc|cccc|cccc}
        \toprule
        & \multicolumn{4}{c}{PSNR $\uparrow$} & \multicolumn{4}{c}{SSIM $\uparrow$} & \multicolumn{4}{c}{LPIPS $\downarrow$} \\
        \cmidrule(r){2-13}
        &EndoNeRF&RoDynRF &EndoGS&Ours&EndoNeRF&RoDynRF &EndoGS&Ours&EndoNeRF&RoDynRF &EndoGS&Ours\\
        \midrule
         pulling soft tissues&\underline{35.832}&27.502&\textbf{36.750}&35.097&\textbf{0.941}&0.900&0.925&\underline{0.936}&\textbf{0.058}&0.158&0.135&\underline{0.059}\\         
         cutting tissues twice&\underline{35.720}&21.105&\textbf{37.372}&35.706&\underline{0.935}&0.700&\textbf{0.964}&0.934&\underline{0.063}&0.280&\textbf{0.050}&0.073\\
         \bottomrule
    \end{tabular}
  \end{adjustbox}
   \begin{flushleft}
Best results are \textbf{bolded}; second best are \underline{underlined}.
\end{flushleft}
\end{table*}

\begin{table*}[ht]
    \centering
    \caption{Quantitative analysis of BASED and EndoGS depth estimation on the Hamlyn Rectified18-2 dataset.}
    \label{tab:depth_results}
    \small
    \begin{adjustbox}{width=0.9\linewidth, center}
    \footnotesize
    \setlength{\arrayrulewidth}{0.01pt}
    \begin{tabular}{l||c|c|c|c|c|c|c}
        \toprule
        & Abs Rel$\downarrow$ & Sq Rel$\downarrow$ & RMSE$\downarrow$ & RMSE Log$\downarrow$ & $\delta < 1.25$ $\uparrow$ & $\delta < 1.25^2$ $\uparrow$ & $\delta < 1.25^3$ $\uparrow$ \\
        \midrule
        EndoGS & 3.834 & 1.149 & 2.938 & 4.225 & 3.356 & 6.908 & 10.930 \\

          BASED & \textbf{1.124} & \textbf{0.228} & \textbf{1.790} & \textbf{1.341} & \textbf{10.227} & \textbf{11.000} & \textbf{11.000}\\ 
        \bottomrule
    \end{tabular}
    \end{adjustbox}
\end{table*}




\subsection{Datasets} \label{sec:data}
We evaluated BASED on the following datasets:

\emph{\textbf{Hamlyn Dataset}~\cite{ye2017self}}: We choose three videos from the Hamlyn dataset of robotic surgery with deformable scenes. Hamlyn dataset contains weak textures, and the scene is often occluded by blood and surgical instruments. Rectified04 contains 158 frames, and it appears to have slight camera motion. The camera in Rectified04 is mostly static, whereas Rectified18 exhibits slightly more camera motion, so two subsets of Rectified18 are chosen from the whole video with 113 frames in each. The frames are sampled at regular intervals (every 10 frames), as the original videos were too large. 
10 percent of the total images were used for testing. 
We use \mbox{\cite{recasens2021endo}} to generate reference depth needed for calculating the estimated depth loss. 


\emph{\textbf{EndoNeRF da Vinci dataset}~\cite{wang2022endonerf}}: We used two datasets from the EndoNeRF paper that are publicly available. They last for 4-8 sec with 15 fps. Each of the videos is taken from a single static viewpoint setting. One of the videos demonstrates soft tissues being drastically pulled, while the other video demonstrates soft tissue cutting with challenging topological changes. As mentioned in the original paper, the depth maps are obtained using STTR-LIGHT~\cite{sttrlight} from stereo images.

\subsection{Metrics} The quality of image rendering is evaluated through commonly used metrics, including  (1) Peak Signal-to-Noise Ratio (PSNR), (2) Learned Perceptual Image Patch Similarity (LPIPS)~\cite{zhang2018unreasonable}, and (3) Structural Similarity Index (SSIM)~\cite{zhang2018unreasonable}. 
We also quantitatively evaluate the 3D reconstruction performance by comparing predicted depth maps with the reference depth (refer to Section \mbox{\ref{sec:data}} for details), using standard metrics for depth evaluation \mbox{\cite{eigen2014depth}}. With this, we are treating the reference as pseudo ground truth. It should be noted that while the quality this pseudo ground truth may be slightly inferior than that of true depth, using it is still a common strategy in endoscopic video analysis as true ground truth can be hard to obtain in a surgical environment \mbox{\cite{schmidt2024tracking}}.

\subsection{Results}

\begin{table*}[h]
    \centering
    \caption{Quantitative analysis of BASED and EndoGS depth estimation on the EndoNeRF "Cutting Tissues Twice" dataset.}
    \label{tab:depth_results2}
    \small
    \begin{adjustbox}{width=0.9\linewidth, center}
    \footnotesize
    \setlength{\arrayrulewidth}{0.01pt}
    \begin{tabular}{l||c|c|c|c|c|c|c}
        \toprule
        & Abs Rel$\downarrow$ & Sq Rel$\downarrow$ & RMSE$\downarrow$ & RMSE Log$\downarrow$ & $\delta < 1.25$ $\uparrow$ & $\delta < 1.25^2$ $\uparrow$ & $\delta < 1.25^3$ $\uparrow$ \\
        \midrule
        EndoGS & \num{4.644e6} & \num{1.643e6} & \textbf{54.342} & 622.943 & 52.230 & 69.472 & 85.756 \\

          BASED & \textbf{47.931} & \textbf{27.107} & 65.810 & \textbf{140.360} & \textbf{91.750} & \textbf{92.898} & \textbf{93.772}\\ 
        \bottomrule
    \end{tabular}
    \end{adjustbox}
\end{table*}

\begin{figure}[h]
\centering
\footnotesize
\begin{tabular}{c@{\hspace{0.2cm}}c@{\hspace{0.1cm}}c@{\hspace{0.4cm}}c@{\hspace{-0.1cm}}c}
\footnotesize \hspace{0mm}Ground Truth & \footnotesize BASED (Ours) & \footnotesize EndoNeRF & \footnotesize RoDynRF & \footnotesize \hspace{6mm} EndoGS
\end{tabular}
\includegraphics[width=1\linewidth, clip=true, trim={0 0 0 0}]{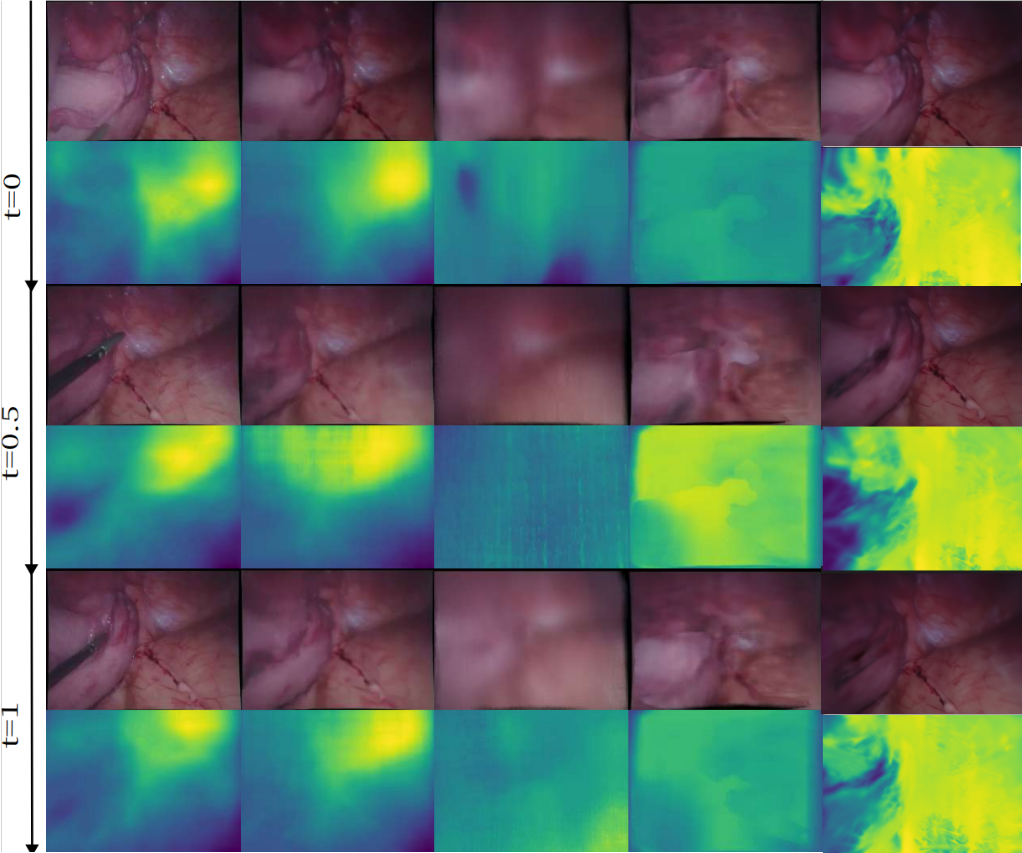}
\captionsetup{font=footnotesize}
\caption{Qualitative analysis of our proposed model BASED compared with RoDynRF, EndoNeRF and EndoGS (with COLMAP estimated poses) for sequences obtained from the public Hamlyn dataset. Please note that the depth in the Ground Truth column is the reference depth.}
\label{fig:hamlynresults}
\end{figure}


\subsubsection{Ablation Study}

Table \ref{tab:rectified18_1_ablation} and Figure~\ref{fig:depth_ablation} provide both quantitative and qualitative analysis of the effect of the various losses. 
As shown in Table~\mbox{\ref{tab:rectified18_1_ablation}}, bundle-adjusting leads to significant improvement in image rendering quality compared with EndoNeRF \mbox{\cite{wang2022endonerf}}. Also, training with estimated depth guidance $L_d$ and correspondence loss $L_{corr}$ improves rendering quality, respectively. 
Figure~\mbox{\ref{fig:depth_ablation}} shows that estimated depth guidance clearly improves the reconstructed textures and depths. 
Additionally, correspondence loss $L_{corr}$ leads to better results, as shown quantitatively by image rendering results and qualitatively by predicted depth quality, indicating that the correspondence loss may have improved the estimated camera poses by instilling multi-view consistency. 
However, the sequences are still severely underconstrained (camera movement/viewing angles of the scene are still limited in those sequences, which is common in surgical scenes), hence the correspondence loss only showed a relatively slight improvement in depth reconstruction. 
The tool masks for sequence - Rectified04-1 from the Hamlyn dataset does not always completely mask out the tools, which is why the resultant rendered images partially have tools in them (Figure~\ref{fig:teaser}), and not a transparent overlay like we see for the other rendered results.



\subsubsection{Comparison on Hamlyn Dataset}

The main dataset we are evaluating is the Hamlyn dataset. We compare BASED with state-of-the-art methods including EndoNeRF, RoDynRF and EndoGS, for deformable scene reconstruction with obvious camera motions. 
The image rendering are quantitatively and qualitatively compared in Table~\mbox{\ref{tab:quantitative_analysis}} and Figure~\mbox{\ref{fig:hamlynresults}}, respectively. 
The 3D scene reconstruction performance is evaluated through predicted depth maps, as shown in Figure \mbox{\ref{fig:hamlynresults}}. 
Our model outperforms both EndoNeRF and RoDynRF by a significant margin in both image rendering and 3D reconstruction.

\begin{figure}[ht]
\centering
\footnotesize
\begin{tabular}
{c@{\hspace{0.2cm}}c@{\hspace{0.1cm}}c@{\hspace{0.4cm}}c@{\hspace{-0.1cm}}c}
\footnotesize \hspace{0mm}Ground Truth & \footnotesize BASED (Ours) & \footnotesize EndoNeRF & \footnotesize RoDynRF & \footnotesize \hspace{6mm} EndoGS
\end{tabular}
\includegraphics[width=1\linewidth]{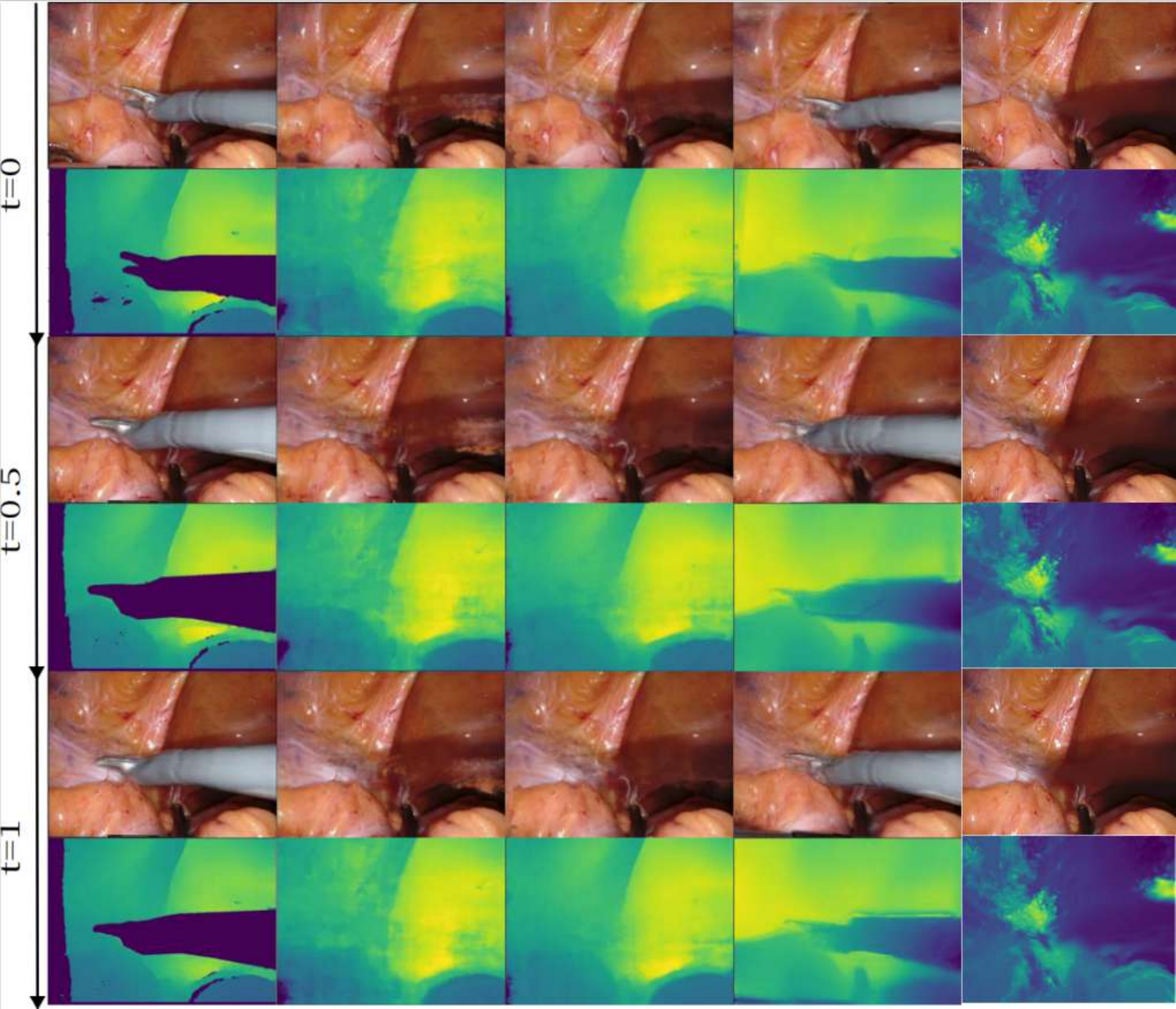}
\captionsetup{font=footnotesize}
\caption{Qualitative analysis of our proposed model BASED with RoDynRF and EndoNeRF for two publicly available sequences - "Cutting Tissues Twice"\cite{wang2022endonerf} guided by stereo depth.}
\label{fig:endonerfresults}
\end{figure}

On the other hand, BASED is still competitive in image rendering compared with EndoGS, which is based on 3DGS. 3DGS has inherent advantages in image rendering, as the color and position of the Gaussians can be easily updated to fit the camera view. However, the depth estimated by EndoGS may not accurately represent the true geometric shape of the environment.
To demonstrate the superiority of our method in capturing the 3D geometric information of the scene, beyond Figure~\mbox{\ref{fig:hamlynresults}}, we also quantitatively measure the quality of predicted depth based on the reference depth, as shown in Table~\mbox{\ref{tab:depth_results}}. BASED provides better depth maps both quantitatively and qualitatively. 
Furthermore, we visualize the camera pose along with the point cloud estimated by EndoGS and BASED in Figure~\mbox{\ref{fig:based model}}. Given the small camera motion in this dataset, BASED aligns more closely with this pattern compared to EndoGS. 
A major drawback of EndoNeRF and EndoGS is that they have only been tested with single-view static cameras, which are severely under-constrained. In dynamic scenes, both models use COLMAP to estimate camera poses.

Thus, these two models fail when COLMAP fails to predict camera poses as seen for Rectified 04-1 in Table~\mbox{\ref{tab:quantitative_analysis}}. This clearly demonstrates the superiority of our method. Also, even in cases where COLMAP is able to estimate camera poses, our model, BASED, more accurately captures the dynamic and deformable aspects of the scene, as shown by the quantitative and qualitative results provided.

\subsubsection{Comparison on EndoNeRF da Vinci Dataset}

We further show that our model can generalize toward deformable surgical scenes with stationary cameras on the sequences: Cutting tissues twice and Pulling soft tissues as shown in Table~\ref{tab:endonerf_reuslts} and Figure~\ref{fig:endonerfresults}. We can see that our model is on par, if not better than EndoNeRF, and performs significantly better than RoDynRF. In Figure~\ref{fig:novel_view}, our model can generalize better novel view renderings compared to EndoNeRF which has been strictly trained with static identity camera poses. Hence, our model retains EndoNeRF's capacity for static deformable monocular videos. 

Furthermore, we still compare our model with EndoGS on static surgical scenes. From Table~\ref{tab:depth_results2} and Figure~\ref{fig:endonerfresults}, both qualitative and quantitative results show although EndoGS shows its inherent advantage in image rendering, it still almost fails in geometric shape capturing even on static camera scenes, which is very crucial for real-world applications.

\section{Conclusion}

We propose BASED, a network that simultaneously estimates dynamic scene motions and reconstructs deformable surgical scenes from monocular videos. This is the first work that tries to reconstruct scenes from video sequences that contain mixed static and dynamic parts with unknown monocular camera poses. Through our experiments, we have shown significant improvement over the existing methods, and also potential for future research.
Limitations of the presented model include slow optimization time (similar to EndoNeRF at several hours), though many recent strategies are drastically cutting this time by orders of magnitude (including ours~\cite{liu2023baangp}). 
Some potential future directions could be removing the dependency of BASED on reference depths, reducing optimization time, and automatic tool mask generation.



{\small
\bibliographystyle{ieee_fullname}
\bibliography{egbib}
}

\end{document}